\documentclass[runningheads]{llncs}
\usepackage{graphicx}
\usepackage{amsmath,amssymb} % define this before the line numbering.
\usepackage{color}
\usepackage[width=122mm,left=12mm,paperwidth=146mm,height=193mm,top=12mm,paperheight=217mm]{geometry}
\usepackage[font=footnotesize]{caption}
\usepackage{graphicx,wrapfig,lipsum}
\usepackage{booktabs}       % professional-quality tables
\usepackage{tabularx}
\usepackage{mwe}

\begin{document}
% \linenumbers
\pagestyle{headings}
\mainmatter
\def\ECCV16SubNumber{***}  % Insert your submission number here

\title{Deep Koalarization: Image Colorization using CNNs and Inception-Resnet-v2 \thanks{Project developed in June 2017}} % Replace with your title

\titlerunning{Image Colorization using CNNs and Inception-ResNet-v2}

\authorrunning{Federico, Diego, Lucas}

\author{Federico Baldassarre$^{**}$, Diego Gonz\'alez Mor\'in$^{**}$, Lucas Rod\'es-Guirao\thanks{Equal contribution} \\ \texttt{\{fedbal, diegogm, lucasrg\} @kth.se }}
\institute{KTH Royal Institute of Technology}

\maketitle
%deep CNN architecture
\begin{abstract}
% Where you give an overview of the task and the findings of your work in a nutshell.
%
% EVALUATION
% Abstract, problem formulation and motivation: 10
% – Language and readability: 3
% – Proper motivation of the problem: 2
% – Accuracy and understandability of the problem formulation: 5

We review some of the most recent approaches to colorize gray-scale images using deep learning methods. Inspired by these, we propose a model which combines a deep Convolutional Neural Network trained from scratch with high-level features extracted from the Inception-ResNet-v2 pre-trained model. Thanks to its fully convolutional architecture, our encoder-decoder model can process images of any size and aspect ratio. Other than presenting the training results, we assess the ``public acceptance" of the generated images by means of a user study. Finally, we present a carousel of applications on different types of images, such as historical photographs. %fashion photographs, art, surveillance camera pictures etc.
\keywords{Deep Learning, Colorization, CNN, Inception-ResNet-v2, Transfer Learning, Keras, TensorFlow}
\end{abstract}

%%%%%%%%%%%%%%%%%%%%%%%%%%%%%%%%%%%%%%%%%%%%%%%%%%%%%%
%%%% 1. INTRODUCTION %%%%%%%%%%%%%%%%%%%%%%%%%%%%%%%%%
%%%%%%%%%%%%%%%%%%%%%%%%%%%%%%%%%%%%%%%%%%%%%%%%%%%%%%
% Motivate the problem you are trying to solve, attempt to make an intuitive description of the problem and also formally define the problem. (1-2 pages including title, authors and abstract)

\section{Introduction}
Coloring gray-scale images can have a big impact in a wide variety of domains, for instance, re-master of historical images and improvement of surveillance feeds. The information content of a gray-scale image is rather limited, thus adding the color components can provide more insights about its semantics. In the context of deep learning, models such as Inception \cite{inception3}, ResNet \cite{resnet} or VGG \cite{vgg16} are usually trained using colored image datasets. When applying these networks on gray-scale images, a prior colorization step can help improve the results. However, designing and implementing an effective and reliable system that automates this process still remains nowadays as a challenging task. The difficulty increases even more if we aim at fooling the human eye. \\

In this regard, we propose a model that is able to colorize images to a certain extent, combining a deep Convolutional Neural Network architecture and the latest released Inception model to this date, namely Inception-ResNet-v2 \cite{inceptionresnetv2}, which is based on Inception v3 \cite{inception3} and Microsoft's ResNet \cite{resnet,resnet2}. While the deep CNN is trained from scratch, Inception-ResNet-v2 is used as a high-level feature extractor which provides information about the image contents that can help their colorization.

Due to time constraints, the size of the training dataset is considerably small, which leads to our model being restricted to a limited variety of images. Nevertheless, our results investigate some approaches carried out by other researchers and validates the possibility to automate the colorization process.

\subsection{Contribution}
In brief, our contribution in this report can be summarized as follows.

\begin{itemize}
\item High-level feature extraction using a pre-trained model (Inception-ResNet-v2) to enhance the coloring process.%Inception-ResNet-v2 to extract high-level features from .
\item Analysis and intuition behind a colorization architecture based on CNNs.
\item Public acceptance evaluation by means of a user study.
\item Test of the model with historical pictures.
\end{itemize}

\subsection{Organization}
Section \ref{sec:2} briefly dives into the origins of image coloring techniques. Section \ref{sec:3} aims at presenting our approach and detailing its main components. Next, Section \ref{sec:4} presents our results, illustrating some colored images, and validates their ``public acceptance" through a user study. Finally, Section \ref{sec:5} concludes the report with some notes on future work.

%Colorization of  %Since the

%TODO: (1) Talk about boom in deep learning, can be used in colorization etc., (2) Comment something of Inception-ResNet-v2 \cite{inceptionresnetv2}...
%IMPORTANT: Carefully think what to put in introduction and what to put in background. See references to see how they split the content!

%Bleeding edge, Deep learning techniques

%%%%%%%%%%%%%%%%%%%%%%%%%%%%%%%%%%%%%%%%%%%%%%%%%%%%%%
%%%% 2. Background %%%%%%%%%%%%%%%%%%%%%%%%%%%%%%%%%%%
%%%%%%%%%%%%%%%%%%%%%%%%%%%%%%%%%%%%%%%%%%%%%%%%%%%%%%
% summarize a few notable approaches/papers tackling the same problem. The selection should cover different possible tech- niques that can be (have been) used for the same task with success. Also, it is good to mention other recognition/synthesis tasks that use the same deep learning technique as yours. (1-2 pages)
%
% EVALUATION
% Background: 15 + 5 as bonus
% – Language and readability: 3
% – Coverage of the background: 7
% – Accuracy of other works representation: 5

\section{Background} \label{sec:2}
In 2002, Welsh \emph{et. al.} \cite{welsh2002transferring} presented a novel approach  which was able to colorize an input image by transferring the color from a related reference image. Subsequent improvements of this method were proposed, exploiting low-level features \cite{ironi2005colorization} and introducing multi-modality on the pixel color values \cite{charpiat2008automatic}. In parallel, another research line was initiated in 2004 by Levin \emph{et. al.}, who proposed a scribble based method \cite{levin2004colorization} which required the user to specify the colors of few image regions. This colorization methodology woke the interest of animators and cartoon-aimed techniques were proposed \cite{qu2006manga,sykora2009lazybrush}. The results from these approaches were certainly impressive at that time, however, the results were highly dependent on the artistic skills of the user. More recently, automatized approaches have been proposed. For instance, in \cite{deshpande2015learning} Desphande \emph{et al.} conceived the coloring problem as a linear system problem.

%It has rained a lot since Hubel and Wiesel \cite{fukushima1988neocognitron} defined an Artificial Neural Network as a hierarchical organization of features.
In the last years, CNNs have been proven experimentally to almost halve the error rate for object recognition \cite{krizhevsky2012imagenet}, which has led to a massive shift towards deep learning of the computer vision community. In this regard, Cheng Z. \emph{et al.} \cite{cheng2015deep} proposed a deep neural network using image descriptors (luminance, DAISY features \cite{tola2008fast} and semantic features) as inputs. In 2016, Iizuka, Serra \emph{et. al.} \cite{iizuka2016let} proposed a method based on using global-level and mid-level features to encode the images and colorize them. Our model draws its architecture on their approach and also serves as a validation. However, we introduce a pre-trained model into the equation. It is worth saying that similar approaches have been presented lately as well. For instance, Zhang \emph{et. al.} \cite{zhang2016colorful} proposed a multi-modal scheme, where each pixel was given a probability value for each possible color. Another interesting approach was developed by Larsson \emph{et al.} \cite{larsson2016learning}, in which a fully convolutional version of VGG-16 \cite{simonyan2015VGG} with the classification layer discarded was used to build a color probability distribution for each pixel.
Recently, Zhang \emph{et al.} \cite{zhang2017real} presented an end-to-end CNN approach incorporating user ``hints" in the spirit of scribble based methods, providing a color recommender system to help novice users and claiming to have enabled real-time use of their colorization system. This recent research proves that this is an ongoing research line.

\section{Approach} \label{sec:3}

We consider images of size $H \times W$ in the CIE L*a*b* color space. \cite{robertson1977cie}. Starting from the luminance component $\textbf{X}_\textbf{L} \in \mathbb{R}^{H\times W \times 1}$, the purpose of our model is to estimate the remaining components to generate a fully colored version $\tilde{\textbf{X}} \in \mathbb{R}^{H \times W \times 3}$. In short, we assume that there is a mapping $\mathcal{F}$ such that

\begin{equation}
\mathcal{F} : \textbf{X}_{\textbf{L}} \to (\tilde{\textbf{X}}_{\textbf{a}}, \tilde{\textbf{X}}_{\textbf{b}}),
\end{equation}

where $\tilde{\textbf{X}}_{\textbf{a}}$, $\tilde{\textbf{X}}_{\textbf{b}}$ are the a*, b* components of the reconstructed image, which combined with the input give the estimated colored image $\tilde{\textbf{X}} = ({\textbf{X}}_{\textbf{L}}, \tilde{\textbf{X}}_{\textbf{a}}, \tilde{\textbf{X}}_{\textbf{b}})$.

In order to be independent of the input size, our architecture is fully based on CNNs, a model that has been extensively studied and employed in the literature \cite{Goodfellow-et-al-2016}. In brief, a convolutional layer is a set of small learnable filters that fit specific local patterns in the input image. Layers close to the input look for simple patterns such as contours, while the ones closer to the output extract more complex features \cite{zeiler2014visualizing}.

%Each layer outputs an array of feature maps, as illustrated in Fig. \ref{fig:conv_layer}.

%scan

As already pointed out, we choose the CIE L*a*b* color space to represent the input images, since it separates the color characteristics from the luminance that contains the main image features \cite{luminance2015}\cite{feature2002}. Combining the luminance with the predicted color components ensures a high level of detail on the final reconstructed image. \\

%Furthermore, CIE L*a*b* has also the property of device independence. The color representation does not depend on the nature of creation ( e.g. different camera devices, processing software... ) or the device they are displayed on. For instance, CIE L*a*b* colorspace is used as an intermediate step in image color conversion, as it guarantees a correct color preservation. This characteristics was also important for the correct implementation of the current model, as it ensure the correct training with independence of the device where the images where load from or the nature of the images themselves. \\

%\begin{itemize}
%\item L* = (0, 100).
%\item a* = (-128,127).
%\item b* = (-128, 127)
%\end{itemize}
\subsection{Architecture}

\begin{figure}[h!]
\centering
\includegraphics[scale=0.18]{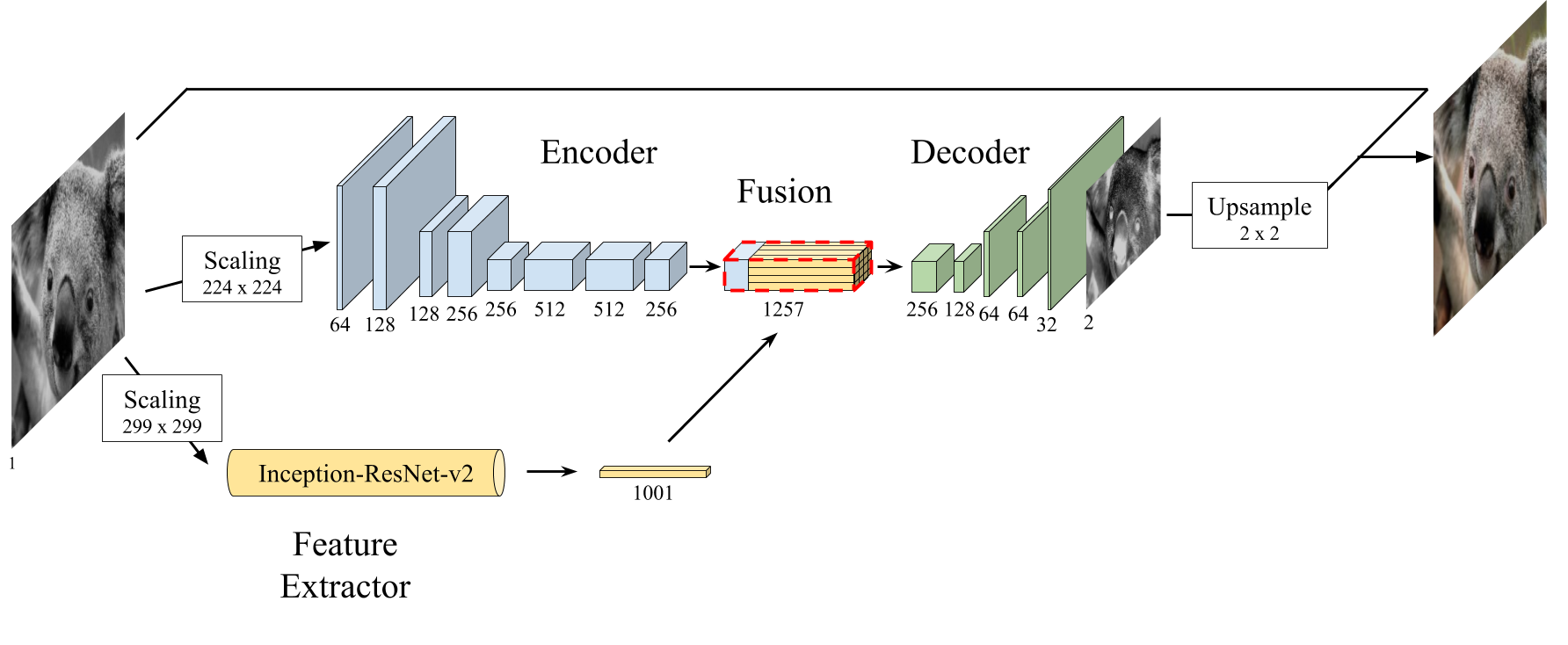}
\caption{An overview of the model architecture.} %The loss function is computed between the a*b* estimates and the a*b* ground truth.}
\label{fig:net}
\end{figure}

Our model owes its architecture to \cite{iizuka2016let}: given the luminance component of an image, the model estimates its a*b* components and combines them with the input to obtain the final estimate of the colored image. Instead of training a feature extraction branch from scratch, we make use of an Inception-ResNet-v2 network (referred to as Inception hereafter) and retrieve an embedding of the gray-scale image from its last layer. The network architecture we propose is illustrated in Fig. \ref{fig:net}.

\newpage

The network is logically divided into four main components. The encoding and the feature extraction components obtain mid and high-level features, respectively, which are then merged in the fusion layer. Finally, the decoder uses these features to estimate the output. Table \ref{tab:arch} further details the network layers.

\begin{table}[h!]
    \caption{ Left: encoder network, mid: fusion network, right: decoder network. Each convolutional layer uses a ReLu activation function, except for the last one that employs a hyperbolic tangent function. The feature extraction branch has the same architecture of Inception-Resnet-v2, excluding the last softmax layer.}
    \vspace*{.1cm}
	\label{tab:arch}
    \scriptsize
    \centering
    \begin{tabular}{ccc}
      \begin{tabular}[t]{lcc|}
        \textbf{Layer} & \textbf{Kernels} & \textbf{Stride}\\
        \hline
        conv & $64 \times (3\times 3)$ & $2\times 2$\\
        conv & $128 \times (3\times 3)$ & $1\times 1$\\
        conv & $128 \times (3\times 3)$ & $2\times 2$\\
        conv & $256 \times (3\times 3)$ & $1\times 1$\\
        conv & $256 \times (3\times 3)$ & $2\times 2$\\
        conv & $512 \times (3\times 3)$ & $1\times 1$\\
        conv & $512 \times (3\times 3)$ & $1\times 1$\\
        conv & $256 \times (3\times 3)$ & $1\times 1$\\
      \end{tabular}
      \begin{tabular}[t]{lcc}
        \textbf{Layer} & \textbf{Kernels} & \textbf{Stride}\\
        \hline
        fusion & - & - \\
        conv & $256 \times (1\times 1)$ & $1\times 1$\\
      \end{tabular}
      \begin{tabular}[t]{|lcc}
        \textbf{Layer} & \textbf{Kernels} & \textbf{Stride}\\
        \hline
        conv & $128 \times (3\times 3)$ & $1\times 1$\\
        upsamp & - & - \\
        conv & $64 \times (3\times 3)$ & $1\times 1$\\
        conv & $64 \times (3\times 3)$ & $1\times 1$\\
        upsamp & - & - \\
        conv & $32 \times (3\times 3)$ & $1\times 1$\\
        conv & $2 \times (3\times 3)$ & $1\times 1$\\
        upsamp & - & - \\
      \end{tabular}
    \end{tabular}
\end{table}

\subsubsection{Preprocessing} To ensure correct learning, the pixel values of all three image components are centered and scaled (according to their respective ranges \cite{hoffman2003cielab}) in order to obtain values within the interval of $[-1, 1]$.

\subsubsection{Encoder} The Encoder processes $H \times W$ gray-scale images and outputs a $H/8 \times W/8 \times 512$ feature representation. To this end, it uses 8 convolutional layers with $3\times 3$ kernels. Padding is used to preserve the layer's input size. Furthermore, the first, third and fifth layers apply a stride of 2, consequentially halving the dimension of their output and hence reducing the number of computations required \cite{stride}.

\subsubsection{Feature Extractor}
\begin{figure}[!ht]
\begin{minipage}{.47\textwidth}
  \centering
\includegraphics[scale=0.075]{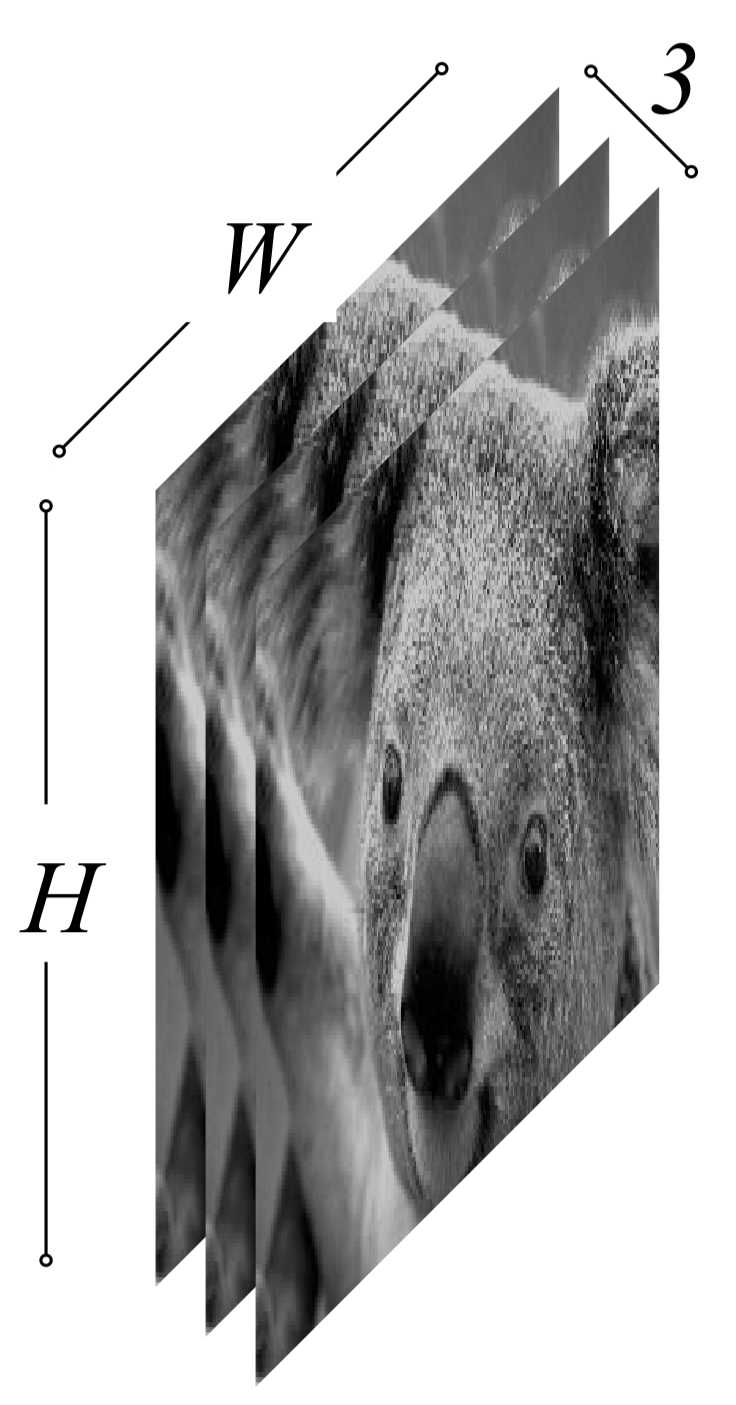}
  \captionof{figure}{Stacking the luminance component three times}
  \label{fig:test1}
\end{minipage}%
\quad
\begin{minipage}{.47\textwidth}
  \centering
\includegraphics[scale=0.08]{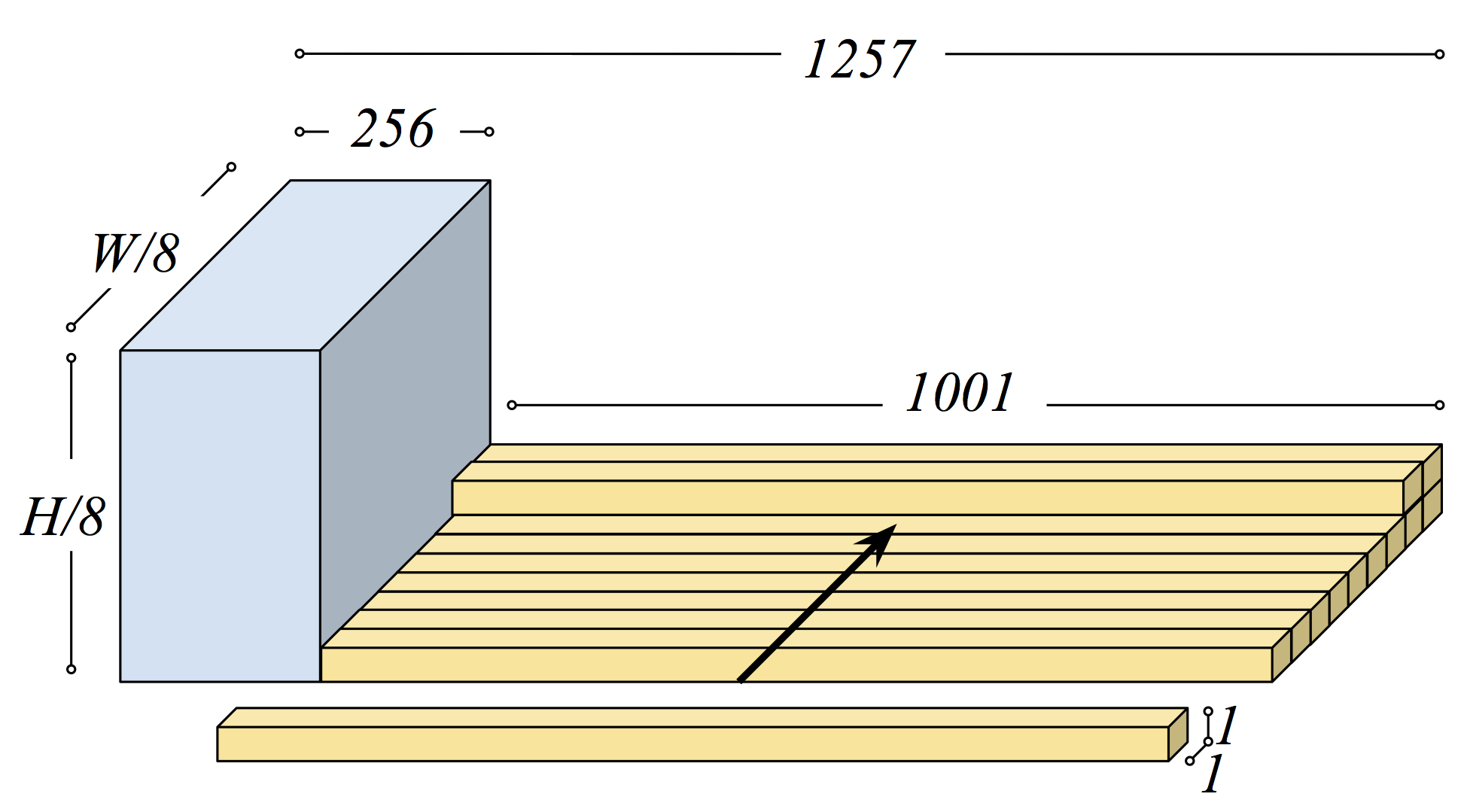}
\vspace*{.65cm}
  \captionof{figure}{Fusing the Inception embedding with the output of the convolutional layers of the encoder}
  \label{fig:test2}
\end{minipage}
\end{figure}

High-level features, e.g. ``underwater" or ``indoor scene", convey image information that can be used in the colorization process. To extract an image embedding we used a pre-trained Inception model. First, we scale the input image to $299 \times 299$. Next, we stack the image with itself to obtain a three-channel image (as shown in Fig. \ref{fig:test1}) in order to satisfy Inception's dimension requirements. Next, we feed the resulting image to the network and extract the output of the last layer before the softmax function. This results in a $1001 \times 1 \times 1$ embedding.

\subsubsection{Fusion}
The fusion layer takes the feature vector from Inception, replicates it $HW/8^2$ times and attaches it to the feature volume outputted by the encoder along the depth axis. This method was introduced by \cite{iizuka2016let} and is illustrated in Fig. \ref{fig:test2}.
This approach obtains a single volume with the encoded image and the mid-level features of shape $H/8 \times H/8 \times 1257$. By mirroring the feature vector and concatenating it several times we ensure that the semantic information conveyed by the feature vector is uniformly distributed among all spatial regions of the image. Moreover, this solution is also robust to arbitrary input image sizes, increasing the model flexibility. Finally, we apply 256 convolutional kernels of size $1 \times 1$, ultimately generating a feature volume of dimension $H/8 \times W/8 \times 256$.

\subsubsection{Decoder} Finally, the decoder takes this $H/8 \times W/8 \times 256$ volume and applies a series of convolutional and up-sampling layers in order to obtain a final layer with dimension $H \times W \times 2$. Up-sampling is performed using basic nearest neighbor approach so that the output's height and width are twice the input's.

\subsection{Objective Function and Training}
The optimal model parameters are found by minimizing an objective function defined over the estimated output and the target output. In order to quantify the model loss, we employ the Mean Square Error between the estimated pixel colors in a*b* space and their real value. For a picture $\textbf{X}$, the MSE is given by (\ref{eq:mse}),

\begin{equation}
C(\textbf{X}, \boldsymbol{\theta}) = \frac{1}{2HW} \sum_{k \in \{a, b\}} \sum_{i=1}^H \sum_{j=1}^W (X_{k_{i,j}} - \tilde{X}_{k_{i,j}})^2,
\label{eq:mse}
\end{equation}

where $\boldsymbol{\theta}$ represents all model parameters, ${X}_{k_{i,j}}$ and $\tilde{X}_{k_{i,j}}$ denote the $ij$:th pixel value of the $k$:th component of the target and reconstructed image, respectively. This can easily be extended to a batch $\mathcal{B}$ by averaging the cost among all images in the batch, i.e. $1/|\mathcal{B}| \sum_{\textbf{X} \in \mathcal{B}} C(\textbf{X}, \boldsymbol{\theta})$.

While training, this loss is back propagated to update the model parameters $\boldsymbol{\theta}$ using Adam Optimizer \cite{adam2014} with an initial learning rate $\eta = 0.001$.
During training, we impose a fixed input image size to allow for batch processing.

%Downside of this approach is that it does not probabilistic, forcing an apple to be always green/red.

%%%%%%%%%%%%%%%%%%%%%%%%%%%%%%%%%%%%%%%%%%%%%%%%%%%%%%
%%%% 4. EXPERIMENTS AND DISCUSSION %%%%%%%%%%%%%%%%%%%
%%%%%%%%%%%%%%%%%%%%%%%%%%%%%%%%%%%%%%%%%%%%%%%%%%%%%%
% In this section, you should present the results you achieved with various experiments. The results can be presented in tables, plots, etc. Explain what conclusions you can draw from these set of experiments? The set of experiments and results reported here should justify some of the design choices described in the previous sections. (3-6 pages)
%
% EVALUATION
% Experiments and conclusions: 23 + 10 as bonus
% – Language and readability: 3
% – Depth and coverage of the experiments: 10
% – Soundness of the evaluation methods: 3
% – Proper use of graphical figures: 3
% – Correctness of the conclusions and discussions: 4
\section{Experiments and Discussion} \label{sec:4}
As important as the network's architecture itself is the choice of the dataset. In the majority of the approaches to automatic image recoloring so far, ImageNet has been extensively used \cite{zhang2016colorful}\cite{larsson2016learning}. Besides, ImageNet's impressive size (more than 14,000,000 images), extensive documentation and free access makes it appealing for our purpose. The dataset is composed of millions of pictures within a wide variety of sets. In particular, it is based on the \textit{name} nodes contained in the word dataset WordNet. In order to simplify training and reduce running times, only a small subset of approximately 60,000 images is used.

ImageNet pictures are heterogeneous in shape, therefore all images in the training set are rescaled to $224 \times 224$ for the encoding branch input and to $299 \times 299$ for Inception. Each image gets stretched or shrunk as needed, but its aspect ratio is preserved by adding a white padding if needed.

\subsection{Training}
Of the approx 60,000 original images, we held out the 10\% to be used as validation data during training. The results presented in this report are drawn from this validation set and therefore the network never had the chance to see those images during training Adam optimizer was used during approximately 23 hours of training.

Complete details about the architecture, the image processing pipeline and our implementation in Keras \cite{chollet2015keras} and TensorFlow \cite{tensorflow2015-whitepaper} can be found in the project webpage\footnote{\small \url{https://github.com/baldassarreFe/deep-koalarization/}}.

The network was trained and tested using the Tegner nodes of The PDC Center for High-Performance Computing at the KTH Royal Institute of Technology, leveraging the NVIDIA® CUDA® Toolkit \cite{Nickolls2008CUDA} and the NVIDIA® Tesla® K80 Accelerator GPU to speed up the computations. A batch size of 100 ruled out the risk of overflowing the GPU memory.

\subsection{Results}
\begin{figure}[h!]
\centering
\includegraphics[width=0.6\textwidth,height=6.3cm]{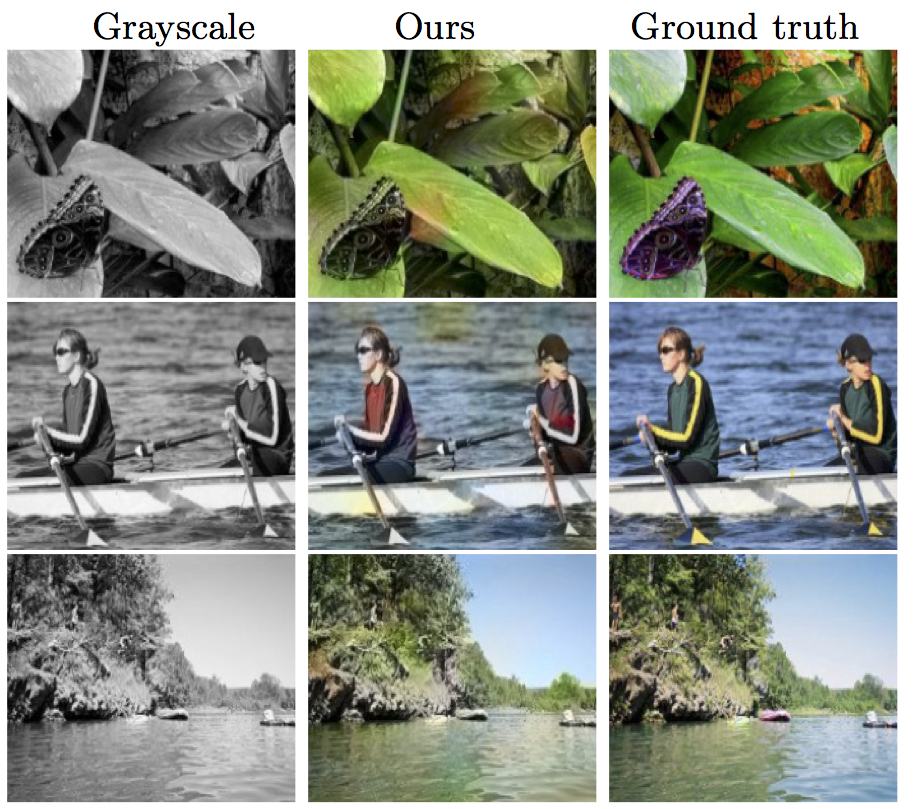}
\caption{In the first row, our approach is capable of recognizing the green vegetation. However, the butterfly was not colored. Furthermore, in the example in the second row, we observe that the network changes the color of the rowers' clothes from green-yellow to red-blue. The last row shows a landscape example where our model provided a photo-realistic image.}
\label{fig:messy}
\end{figure}

Once trained, we fed our network with some images. The results turned out to be quite good for some of the images, generating near-photorealistic pictures. However, due to the small size of our training set our network performs better when certain image features appear. For instance, natural elements such as the sea or vegetation seem to be well recognized. However, specific objects are not always well colored. Fig. \ref{fig:messy} illustrates results for some examples where our network produces alternative colored estimates. %the results of our approach for two pictures. In these example we can observe some irregularities in the colors compared to the ground truth examples.

\begin{figure}[h!]
\centering
\includegraphics[width=1\textwidth,height=12.5cm]{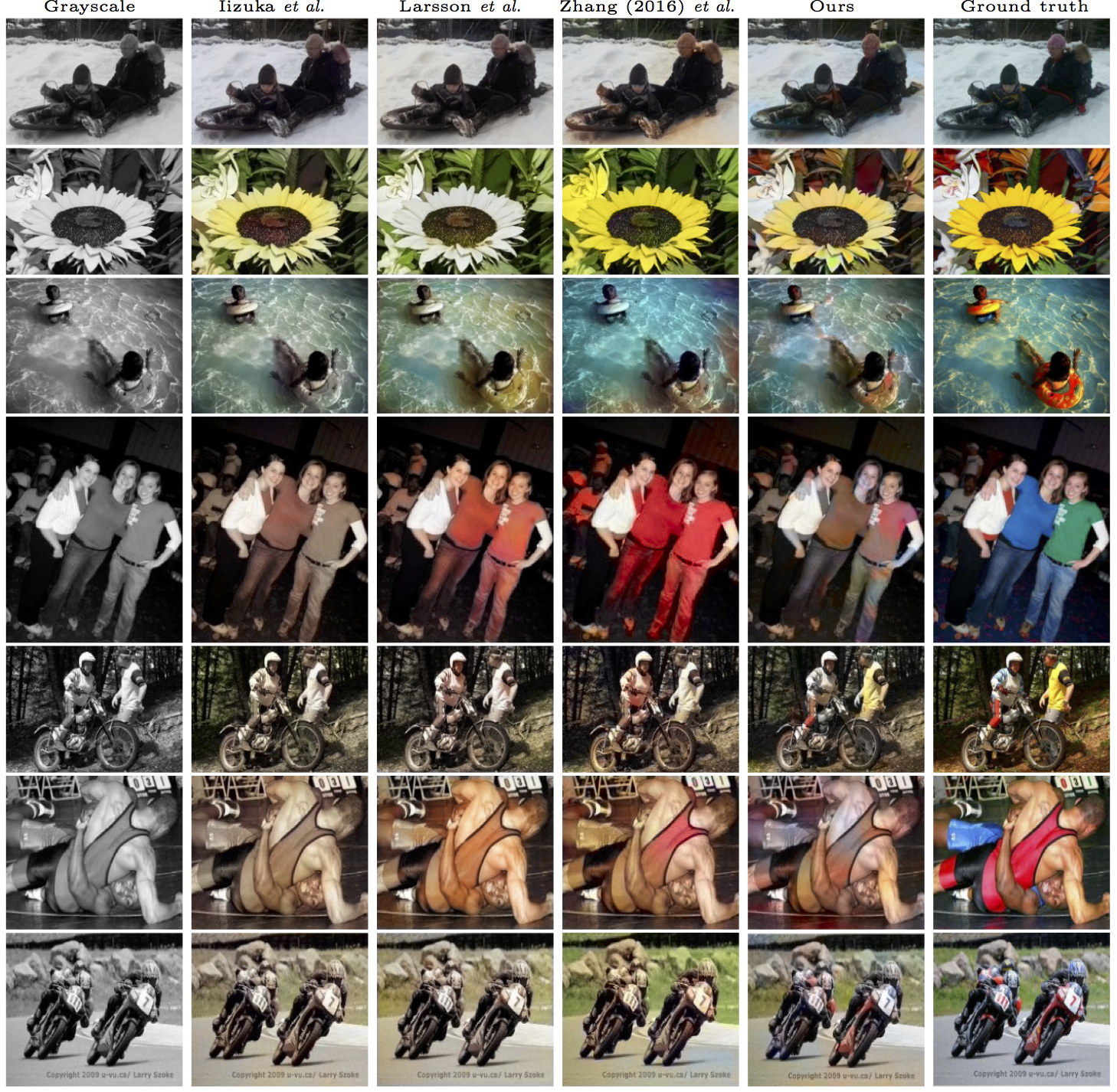}
\caption{Comparison of the results obtained from our colorization network with other approaches. The first column shows the gray-scale input image. Columns 2-4 show the results of the automatic colorization models from Iizuka \emph{et al.}, Larsson \emph{et al.} and Zhang \emph{et al.} (2016), respectively. Column 5 shows our results and, finally, the last column provides the corresponding ground truth images. In the presented examples, in rows 5 and 7 our method outperforms the other methods, generating more photo-realistic images. In the remaining images, some regions of the generated images by our method lack of saturation. Images are from the ImageNet dataset (Russakovsky et al. 2015).}
\label{fig:compare}
\end{figure}

Fig. \ref{fig:compare} exposes generated color images using our method along with other state-of-the-art approaches. Larsson \emph{et al.}, Zhang \emph{et al.} and we used ImageNet training set. Iizuka \emph{et al.}, instead, used Places training dataset. Furthermore, we use the same objective function as Iizuka \emph{et al.} (MSE loss). On the contrary, Larsson \emph{et al.} and Zhang \emph{et al.} use an un-rebalanced and rebalanced classification loss, respectively. From the results, we observed that although some results were quite good, some generated pictures tend to be low saturated, with the network producing a grayish color where the original would be brighter (e.g. with images of fruit, flowers or clothes). Our interpretation is that the network, in its attempt to minimize the loss between images where e.g. flowers are red and others where flowers are blue, ends up doing very doing conservative predictions, namely assigning a neutral gray color.

\subsection{User Study}
\begin{figure}[h!]
\centering
\includegraphics[width=0.55\textwidth,height=6.3cm]{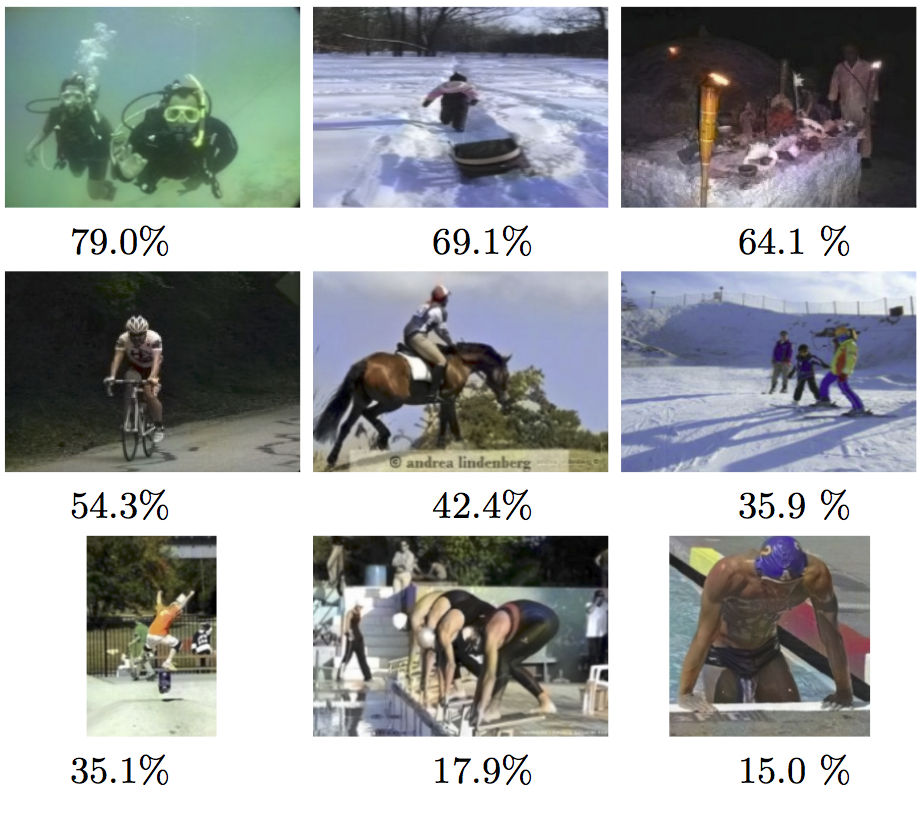}
\caption{For each recolored image we give the percentage of users that answered ``real" to the question \emph{Fake or real?} The images are sorted according to their ``fooling capacity".}
\label{fig:userStudy}
\end{figure}

Although we can empirically obtain a measure of the performance of our model by using (\ref{eq:mse}), we are also interested in how compelling the colors look to a human observer, which can be difficult to assess using solely mathematical tools. Thus, we decided to evaluate the appearance of some artificially recolored images by means of a user study\footnote{\url{https://goo.gl/forms/nxPJUXhmZkeLYmsQ2}}. To this end, we chose twelve images, nine of which were recolored and are shown in Fig. \ref{fig:userStudy}, and asked, for each of them, the question ```Fake or real?". We picked results that we believed could ``fool" the human eye, discarding all the images that were poorly recolored. The poll was taken by 41 different users.

%79.0, 69.2, 64.1, 54.3, 42.4, 35.9, 35.1, 17.9, 15.0
These results totally overcame our expectations. In particular, we can observe that in some cases the real-perception achieved almost 80.0 \%. However, this results also indicate that credibility of the image strongly depends on what the image is portraying. We believe that incrementing the size and variability of the training set could partially mitigate this. Overall, we computed that 45.87\% of the users miss-classified recolored images as originals. However, we have to bear in mind that the recolored images for the user study were carefully selected from our best results.

%shows that 4 out of 9 recolored images were misclassified as Original. In all cases but in Image 2, >60\% of the users identified the recolored images as original. This results are similar to the ones obtained in \cite{zhang2016colorful}. However, in this particular case, the feature extraction was not used. Using all the data obtained in the poll we calculated that 43.73\% of the users classified a recolored image as an original one.

\subsection{Historical Photographs}
\begin{figure}[h!]
\centering
\includegraphics[width=1\textwidth,height=4.2cm]{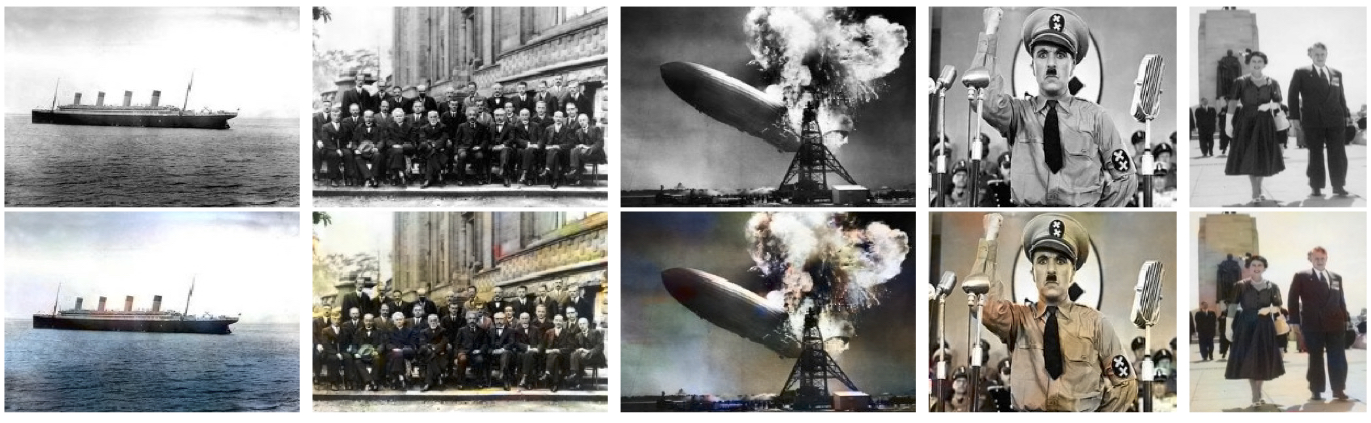}
\caption{ Example of recolored historical images, from left to right: Titanic, before the iceberg(1912),The 1927 Solvay Conference in Brussels(1927), Hindenburg disaster (1937), The Great Dictator (Chaplin, 1940), Queen Elizabeth(1969)}
\label{fig:histPic}
\end{figure}

We tested our model on historical pictures. The results are shown in Fig. \ref{fig:histPic}. Since no ground truth exists, the results are only interpretable by means of personal judgment.

%%%%%%%%%%%%%%%%%%%%%%%%%%%%%%%%%%%%%%%%%%%%%%%%%%%%%%
%%%% 5. FUTURE WORK %%%%%%%%%%%%%%%%%%%%%%%%%%%%%%%%%%
%%%%%%%%%%%%%%%%%%%%%%%%%%%%%%%%%%%%%%%%%%%%%%%%%%%%%%
\section{Conclusions and Future Work} \label{sec:5}
This project validates that an end-to-end deep learning architecture could be suitable for some image colorization tasks. In particular, our approach is able to successfully color high-level image components such as the sky, the sea or forests. Nevertheless, the performance in coloring small details is still to be improved. As we only used a reduced subset of ImageNet, only a small portion of the spectrum of possible subjects is represented, therefore, the performance on unseen images highly depends on their specific contents. To overcome this issue, our network should be trained over a larger training dataset.

In this regard, a probabilistic approach in the spirit of \cite{zhang2016colorful} seems more adequate. We believe that a better mapping between luminance and a*b* components could be achieved by an approach similar to variational autoencoders, which could also allow for image generation by sampling from a probability distribution.

Finally, it could be interesting to apply colorization techniques to video sequences, which could potentially re-master old documentaries. This, of course, would require adapting the network architecture to accommodate temporal coherence between subsequent frames.

Overall, we believe that while image colorization might require some degree of human intervention it still has a huge potential in the future and could eventually reduce hours of supervised work.
\pagebreak

\bibliographystyle{splncs}
\bibliography{paper}
\end{document}